\documentclass[10pt,a4paper,english]{article}\usepackage[]{graphicx}\usepackage[]{xcolor}
\makeatletter
\def\maxwidth{ %
  \ifdim\Gin@nat@width>\linewidth
    \linewidth
  \else
    \Gin@nat@width
  \fi
}
\makeatother

\definecolor{fgcolor}{rgb}{0.345, 0.345, 0.345}

\usepackage{framed}
\makeatletter
 {\par\unskip\endMakeFramed%
 \at@end@of@kframe}
\makeatother

\definecolor{shadecolor}{rgb}{.97, .97, .97}
\definecolor{messagecolor}{rgb}{0, 0, 0}
\definecolor{warningcolor}{rgb}{1, 0, 1}
\definecolor{errorcolor}{rgb}{1, 0, 0}
\newenvironment{knitrout}{}{} 

\usepackage{alltt}

\usepackage[hyphens]{url}
\usepackage{amsmath}
\usepackage{amsfonts}
\usepackage{amssymb}
\usepackage{amsthm}

\theoremstyle{plain}
\newtheorem{theorem}{Theorem}[section]
\newtheorem{lemma}[theorem]{Lemma}

\theoremstyle{definition}

\theoremstyle{remark}

\PassOptionsToPackage{hyphens}{url}\usepackage{hyperref}
\usepackage{hyperref}
\usepackage[square,numbers]{natbib}

\usepackage[newitem,newenum,increaseonly]{paralist}

\usepackage[colorinlistoftodos]{todonotes}


\usepackage[notheorems]{common}

\providecommand{\algorithmref}[1]{Algorithm\nobreakspace\ref{algorithm:#1}}

\providecommand{\YM}[1][{}]{\MtxUL{Y}{#1}{}}
\providecommand{\XM}[1][{}]{\MtxUL{X}{#1}{}}
\providecommand{\XMp}[1][{}]{\XM[\wrapNeParens{#1}]}
\providecommand{\LM}[1][{}]{\MtxUL{L}{#1}{}}
\providecommand{\LMp}[1][{}]{\LM[\wrapNeParens{#1}]}
\providecommand{\RM}[1][{}]{\MtxUL{R}{#1}{}}
\providecommand{\RMp}[1][{}]{\RM[\wrapNeParens{#1}]}
\providecommand{\trLMp}[1][{}]{\tr{\LMp[#1]}}
\providecommand{\trRMp}[1][{}]{\tr{\RMp[#1]}}
\providecommand{\gets}{\leftarrow}

\providecommand{\HM}[1][{}]{\MtxUL{H}{#1}{}}
\providecommand{\HMp}[1][{}]{\HM[\wrapNeParens{#1}]}
\providecommand{\KM}[1][{}]{\MtxUL{K}{#1}{}}
\providecommand{\KMp}[1][{}]{\KM[\wrapNeParens{#1}]}

\providecommand{\DM}[1][{}]{\MtxUL{D}{#1}{}}

\providecommand{\FM}[1][{}]{\MtxUL{F}{#1}{}}

\providecommand{\WM}[2][{}]{\MtxUL{W}{#1}{#2}}

\providecommand{\GM}[1][{}]{\MtxUL{G}{#1}{}}
\providecommand{\GMp}[1][{}]{\MtxUL{G}{\wrapNeParens{#1}}{}}
\providecommand{\dv}[1][{}]{\vectUL{d}{#1}{}}
\providecommand{\dvp}[1][{}]{\vectUL{d}{\wrapNeParens{#1}}{}}
\providecommand{\bv}[1][{}]{\vectUL{b}{#1}{}}

\providecommand{\xmp}[1][{}]{\vectUL{x}{\wrapNeParens{#1}}{}}
\providecommand{\hstep}[1][{}]{\vectUL{h}{\wrapNeParens{#1}}{}}
\providecommand{\Hstep}[1][{}]{\HM[\wrapNeParens{#1}]}
\providecommand{\problemref}[1]{Problem\nobreakspace\ref{eqn:#1}}
\providecommand{\eqnsref}[1]{Equations\nobreakspace\ref{eqn:#1}}

\providecommand{\hadt}{\MATHIT{\odot}}
\providecommand{\hadd}{\MATHIT{\oslash}}

\providecommand{\objf}[2][{}]{\funcitUL{\phi}{#1}{}{#2}}
\providecommand{\objfp}[2][{}]{\funcitUL{\phi}{\wrapNeParens{#1}}{}{#2}}

\providecommand{\AkronB}[2]{\MATHIT{{#1} \kron {#2}}}
\providecommand{\BtkronA}[2]{\AkronB{\tr{#2}}{#1}}
\providecommand{\normsq}[2][2]{\normUL{#1}{2}{#2}}

\providecommand{\mone}[1][]{\MtxUL{1}{}{#1}}
\providecommand{\offdiag}[1][]{\vone\tr{\vone} - \eye[{#1}]}

    \usepackage{enumerate} 
		\usepackage{fancyvrb} 
    \definecolor{orange}{cmyk}{0,0.4,0.8,0.2}
    \definecolor{darkorange}{rgb}{.71,0.21,0.01}
    \definecolor{darkgreen}{rgb}{.12,.54,.11}
    \hypersetup{
      breaklinks=true,  
      colorlinks=true,
      urlcolor=blue,
      linkcolor=darkorange,
      citecolor=darkgreen,
      }

		\let\oldciteauthor=\citeauthor
		\def\citeauthor#1{\hypersetup{citecolor=black}\oldciteauthor{#1}}
    
\usepackage{algorithm}
\usepackage{algpseudocode}

\IfFileExists{upquote.sty}{\usepackage{upquote}}{}
\begin{document}

\title{An Iterative Algorithm for Regularized Non-negative Matrix Factorizations}
\author{Steven E. Pav \thanks{\email{steven@gilgamath.com}
Application of the multiplicative methods described herein for chemometrics applications may be covered by U.S. Patent.  \cite{pav_nmf}
}}

\maketitle

\begin{abstract}
We generalize the non-negative matrix factorization algorithm of
\citeauthor{LeeSeung} \cite{LeeSeung} to accept
a weighted norm, and to support ridge and Lasso regularization.
We recast the \citeauthor{LeeSeung} multiplicative update
as an additive update which does not get stuck on zero values.
We apply the companion R package \textsc{rnnmf} \cite{rnnmfrnnmf-Manual}
to the problem of finding a reduced rank representation of a database
of cocktails.
\end{abstract}

\nocite{LeeSeung,GonzalezZhang,pav_nmf,2024survey}

\section{Introduction}

The term \emph{non-negative matrix factorization} refers to decomposing a 
given matrix \YM with non-negative elements approximately as the product 
$\YM \approx \LM\RM$, for comformable non-negative matrices \LM and \RM of smaller
rank than \YM.
This factorization represents a kind of dimensionality reduction.
The quality of the approximation is measured by some ``divergence'' between
\YM and $\LM\RM$, or some norm of their difference.
Perhaps the most common is the Frobenius (or elementwise $\ell_2$) distance
between them.
Under this objective, non-negative matrix factorization can be expressed as 
the optimization problem:
\begin{equation}
	\min_{\LM \ge 0,\,\RM \ge 0} \trace{\gram{\wrapParens{\YM - \LM\RM}}},
	\label{eqn:nmf_def_classic}
\end{equation}
where $\trace{\Mtx{M}}$ is the trace of the matrix $\Mtx{M}$, and $\Mtx{A} \ge 0$ means
the elements of $\Mtx{A}$ are non-negative.
\cite{nmf2024survey}

\citeauthor{LeeSeung} 
described an iterative approach to this problem, as well as for a related optimization with a different objective. \citep{LeeSeung}
Their algorithm starts with some unspecified initial iterates \LMp[0] and \RMp[0], 
then iteratively improves \LMp[k] and \RMp[k] in turn. That is, given \LMp[k] and \RMp[k],
an update is made to arrive at \LMp[k+1]; then using \LMp[k+1] and \RMp[k], one computes
\RMp[k+1], and the process repeats.  The iterative update for \LMp[k+1] is
\begin{equation}
	\LMp[k+1] \leftarrow \LMp[k] \hadt \YM\tr{\RMp[k]} \hadd \wrapParens{\LMp[k]\ogram{\RMp[k]}}.
	\label{eqn:nmf_def_update}
\end{equation}
We use \hadt to mean the Hadamard, or `elementwise', multiplication, and
$\hadd$ to mean Hadamard division.
The iterative update for \RMp[k+1] is defined \emph{mutatis mutandis}.
As should be apparent from inspection, this iterative update maintains non-negativity of estimates. 
That is, if \YM, \LMp[k], and \RMp[k] have only positive elements, then 
\LMp[k+1] is finite and has only positive elements.
It is not clear what should be done if elements of the denominator term
\LMp[k]\ogram{\RMp[k]} are zero, and whether and how this can be avoided.

\citeauthor{GonzalezZhang} describe a modification of the update to accelerate convergence.
\citep{GonzalezZhang}
Their accelerated algorithm updates \LMp[k] row by row, 
and \RMp[k] column by column,
using the step length modification of 
\citet{merritt2005interior}.

One attractive feature of the 
\citeauthor{LeeSeung} 
iterative update is its sheer simplicity.  
In a high-level computer language with a rich set of matrix operations, 
such as Julia or octave,
the iterative step can be expressed in a single line of code; the entire
algorithm could be expressed in perhaps 20 lines of code. 
Conceivably this algorithm could even be implemented in challenged computing environments, 
like within a SQL query or a spreadsheet macro language.
Simplicity makes this algorithm an obvious choice for experimentation without requiring
an all-purpose constrained quadratic optimization solver, or other complicated software.


\section{Regularized Non-Negative Matrix Factorization}

We generalize the optimization problem of \ref{eqn:nmf_def_classic} to include
a weighted norm, regularization terms, and orthogonality penalties.
For a given \YM, and non-negative weighting matrices
\WM{0,R}, \WM{0,C}, \WM{1,L}, \WM{1,R}, \WM{2,R,L,j}, \WM{2,C,L,j}, \WM{2,R,R,j}, and \WM{2,C,R,j}, find
matrices \LM and \RM to minimize
\begin{align}
\label{eqn:nmf_def_regularized}
\min_{\LM \ge 0,\,\RM \ge 0} \,\,& \half \trace{\qform{\WM{0,R}}{\wrapParens{\YM - \LM\RM}}\WM{0,C}}\\
	\nonumber&\phantom{=}\, 
	+ \trace{\trAB{\WM{1,L}}{\LM}}
	+ \trace{\trAB{\WM{1,R}}{\RM}}\\
	\nonumber&\phantom{=}\, 
	+ \half\sum_{1 \le j \le J}\trace{\qform{\WM{2,R,L,j}}{\LM}\WM{2,C,L,j}}\\
	\nonumber&\phantom{=}\, 
	+ \half\sum_{1 \le j \le J}\trace{\qform{\WM{2,R,R,j}}{\RM}\WM{2,C,R,j}}.
\end{align}
Essentially we are trying to model \YM as $\LM\RM$. 
Our loss function is quadratic in the error $\YM -\LM\RM$, 
with weights \WM{0,R} and \WM{0,C}, 
and $\ell_1$ and $\ell_2$ regularization via
the terms with $\WM{1}$ and \WM{2,R,j} and \WM{2,C,j}.  
This optimization problem includes that of 
\ref{eqn:nmf_def_classic} as a special case.

We note that a non-orthogonality penalty term on the \XM can be expressed as
\begin{equation*}
	\frac{c}{2}\trace{\qform{\eye}{\XM}\wrapParens{\offdiag}}.
\end{equation*}
Thus a non-orthogonality penalty term can be expressed with the $\WM{2,R,j}$ and $\WM{2,C,j}$.
Indeed the motivation for including a sum of the $\ell_2$ terms is to allow for
both straightforward regularization, where $\WM{2,R,j} = \eye$ and $\WM{2,C,j}=c \eye$,
and orthogonality terms.
For simplicity of notation, 
in the discussion that follows we will often treat the $\ell_2$ terms as if there were only one of them,
\ie $J=1$, and omit the $j$ subscripts.
Restoring the more general case consists only of expanding to a full sum of $J$ terms.

This problem formulation is appropriate for power users who can meaningfully select the various weighting matrices.
And while it is easy to conceive of a use for general forms for \WM{0,R} and \WM{0,C} (the rows of \YM have different importances,
or the columns of \YM are observed with different accuracies, for example),
most users would probably prefer simpler formulations for the $\WM{1,\cdot}$ and $\WM{2,\cdot,\cdot,j}$ matrices.
To achieve something like an elasticnet factorization with non-orthogonality penalties, one should set $J=2$ and take
\begin{align}
	\nonumber
	\WM{1,L} &= \lambda_{1,L} \mone,&\quad \WM{1,R} &= \lambda_{1,R} \mone,\\
	\nonumber
	\WM{2,R,L,1} &= \lambda_{2,L} \eye,&\quad \WM{2,C,L,1} &= \eye,\\
	\label{eqn:simple_weighting}
	\WM{2,R,R,1} &= \lambda_{2,R} \eye,&\quad \WM{2,C,R,1} &= \eye,\\
	\nonumber
	\WM{2,R,L,2} &= \eye,&\quad \WM{2,C,L,2} &= \gamma_{2,L} \wrapParens{\offdiag},\\
	\nonumber
	\WM{2,R,R,2} &= \eye,&\quad \WM{2,C,R,2} &= \gamma_{2,R} \wrapParens{\offdiag},
\end{align}
where here $\mone$ stands in for an all-ones matrix of the appropriate size. 
All the $\lambda$ and $\gamma$ parameters must be non-negative, 
and the identity matrices are all appropriately sized.
This is the reduced form of the problem that depends mostly on a few scalars.
A simple form for the weighted error is to take
\WM{0,R} and \WM{0,C} to be diagonal matrices,

Depending on the regularization terms, the solution to 
the problem of \ref{eqn:nmf_def_regularized} is unlikely to be unique.
Certain the solution to \ref{eqn:nmf_def_classic} is not unique, since if \Mtx{Q} is a permutation
matrix of the appropriate size, then $\LM\Mtx{Q}$ and $\minv{\Mtx{Q}}\RM$ are another set of
non-negative matrices with the same objective value.
Perhaps the regularization terms can help avoid these kinds of ambiguity in the solution.

\subsubsection{Split Form}

To approximately solve this problem our algorithm starts with some initial estimates of \LM and \RM and 
iteratively alternates between updating the estimate of \LM with the estimate of \RM fixed,
and updating the estimate of \RM with the estimate of \LM fixed.
To discuss these half-step updates, we can describe both of them in a single unified formulation.
So we consider the optimization problem
\begin{equation}
	\min_{\XM \ge 0} \objf{\XM},
	\label{eqn:nmf_def_new}
\end{equation}
where we define the objective function
\begin{align}
\label{eqn:nmf_new_objective}
\objf{\XM} &=\half \trace{\qform{\WM{0,R}}{\wrapParens{\YM - \LM\XM\RM}}\WM{0,C}}\\
	\nonumber&\phantom{=}\, 
	+ \trace{\trAB{\WM{1}}{\XM}}
	+ \half\sum_{1 \le j \le J} \trace{\qform{\WM{2,R,j}}{\XM}\WM{2,C,j}}.
\end{align}
We must assume that \LM has full column rank and \RM has full row rank,
\WM{0,R}, \WM{0,C}, \WM{2,R,j} and \WM{2,C,j} are square, symmetric and positive semidefinite.
The \WM{0,R} has the same number of rows as \YM, and \WM{0,C} has the same number of columns as \YM.
\WM{1} is the same size as $\XM$; we will make further restrictions on \WM{1} below.
The matrices \WM{2,R,j} have the same number of rows as \XM, and \WM{2,C,j} have the same number of columns as \XM.
All the $\WM{i,\cdot}$ are assumed to have non-negative elements, that is $\WM{i,\cdot} \ge 0$.

Our algorithm now consists of filling in an identity for \LM, and estimate of \LM for \XM, and all the \LM-appropriate
weighting matrices into \eqnref{nmf_new_objective} and taking a step to solve \problemref{nmf_def_regularized},
getting an update of \LM.
Then we perform a similar operation to get a new estimate of \RM.


\subsection{Factorization in Vector Form}

The analysis is simplified if we express the problem as one of finding
a vector unknown, so we will `vectorize' the problem.
For matrix \Mtx{M} let \fvec{\Mtx{M}} be the vector which consists of the columns of \Mtx{M}
stacked on top of each other;
let \AkronB{\Mtx{A}}{\Mtx{B}} denote the Kronecker product of $\Mtx{A}$ and $\Mtx{B}$.
\cite{schacke2013kronecker}
See \secref{matrix_identities} in the appendix for some helpful identities involving the Kronecker product, matrix trace, 
and vectorization.

We write \objf{\XM} as a quadratic function of the vector \fvec{\XM}
\begin{equation}
	\begin{split}
		\objf{\XM} &=\half \normsq[\AkronB{\WM{0,C}}{\WM{0,R}}]{\fvec{\YM} -
	 \wrapParens{\BtkronA{\LM}{\RM}}\fvec{\XM}}\\
	&+ \half\sum_j\normsq[\AkronB{\WM{2,C,j}}{\WM{2,R,j}}]{\fvec{\XM}} + \trAB{\fvec{\WM{1}}}{\fvec{\XM}}\\
	&=\half \qform{\GM}{\fvec{\XM}} + \trAB{\dv}{\fvec{\XM}} + c,
	\end{split}
\end{equation}
where
\begin{equation}
	\begin{split}
		\GM &= \qform{\wrapParens{\AkronB{\WM{0,C}}{\WM{0,R}}}}{\wrapParens{\BtkronA{\LM}{\RM}}} + \sum_j\AkronB{\WM{2,C,j}}{\WM{2,R,j}},\\
		&= \wrapParens{\qoform{\WM{0,C}}{\RM}}\kron\wrapParens{\qform{\WM{0,R}}{\LM}} + \sum_j\AkronB{\WM{2,C,j}}{\WM{2,R,j}}.\\
		\dv &= {\fvec{\WM{1}}} -
		\trAB{\wrapParens{\BtkronA{\LM}{\RM}}}{%
			\wrapParens{\AkronB{\WM{0,C}}{\WM{0,R}}}
			\fvec{\YM}},\\
		&= {\fvec{\WM{1}}} -
		\fvec{\tr{\LM}\WM{0,R}\YM\WM{0,C}\tr{\RM}}.
	\end{split}
\end{equation}
We write $\normsq[\Mtx{A}]{\vect{x}}$ to mean $\qform{\Mtx{A}}{\vect{x}}$.
Note that $\GM \ge 0$, but $\dv$ will likely have negative elements. 
In fact, to use Lee and Seung's iterative algorithm, the vector \dv must have non-positive elements. 
This is equivalent to the restriction
$\tr{\LM}{\WM{0,R}}\YM\WM{0,C}\RM - \WM{1} \ge 0$, taken elementwise.
This imposes a restriction on $\WM{1}$ which is hard to verify for the problem of non-negative matrix factorization.
Note that by definition \GM can be positive definite, depending on the \RM, \LM and the $\WM{2,\cdot}$, but is 
only positive \emph{semi} definite in the general case.

We can thus consider this problem as optimization over a vector unknown.
In an abuse of notation, we reuse the objective function \objf{\cdot}, writing
\begin{equation}
\objf{\xmp} = \half \qform{\GM}{\xmp} + \trAB{\dv}{\xmp},
\end{equation}
for symmetric, nonnegative positive semidefinite \GM. 
So now we consider 
the positivity-constrained quadratic optimization problem
\begin{equation}
	\min_{\xmp \ge 0} \objf{\xmp}.\label{eqn:cons_quad_opt_problem}
\end{equation}
Note that the unconstrained solution
to this problem is $-\minv{\GM}{\dv}$.  
When \GM is positive definite, because the constraint space is convex,
this solution will be unique. \cite{nocedal2006numerical}
In general, however, we will not have unique solutions.
When the \WM{1} term is all zero, 
this a weighted least squares problem, possibly with Tikhonov regularization.

The gradient of the objective \objf{\xmp} is 
is $\gradof{\objf{\xmp}} = \GM\xmp + \dv$.
When considering the matrix form of the problem, this has the value
$$
\fvec{\wrapBracks{\qform{\WM{0,R}}{\LM}\XM\qoform{\WM{0,C}}{\RM} + \sum_j\WM{2,R,j}\XM\WM{2,C,j}  + \WM{1} - \tr{\LM}\WM{0,R}\YM\WM{0,C}\tr{\RM}}}.
$$

We consider an iterative solution to this problem. We first select
some \xmp[0]. 
Then given current iterate \xmp[k], we seek to find the next
iterate \xmp[k+1]. 
Define
$$
\objfp[k]{\xmp} = \half\qform{\GMp[k]}{\xmp} + 
\trAB{\dvp[k]}{\xmp} + c_k,
$$
where we will set \GMp[k], \dvp[k] and $c_k$ such that
\objfp[k]{\xmp} is tangent to \objf{\xmp} at \xmp[k], and
such that $\objfp[k]{\xmp} \ge \objf{\xmp}$ for all \xmp.
We will take \xmp[k+1] as some point which improves
\objfp[k]{\xmp}: its minimizer if it is non-negative, or
another point if not.
If \xmp[k+1] improves \objfp[k]{\xmp}, then it also improves
\objf{\xmp}, as
\begin{equation*}
	\objf{\xmp[k+1]} \le \objfp[k]{\xmp[k+1]} \le \objfp[k]{\xmp[k]} = \objf{\xmp[k]}.
\end{equation*}

The tangency condition fixes the value of $c_k$, and also
gives the identity
$$
\dvp[k] = \dvp + \GM\xmp[k] - \GMp[k]\xmp[k].
$$
To ensure that $\objfp[k]{\xmp} \ge \objf{\xmp}$ 
it is sufficient to select select \GMp[k] such that 
$\GMp[k] - \GM$ is positive semidefinite. 
The following lemma
gives us a way of selecting diagonal matrices \GMp[k] that 
satisfy this condition.

First a bit of notation, for vector $\vect{v}$, $\Mdiag{\vect{v}}$ is the diagonal
matrix with $\vect{v}$ for diagonal. 
For matrix $\Mtx{A}$, $\vdiag{\Mtx{A}}$ is the vector of the diagonal of \Mtx{A}.
Thus we have $\vdiag{\Mdiag{\vect{v}}} = \vect{v}$.

\begin{lemma}
	\label{lemma:diag_dominance}
Let \GM be a symmetric matrix with non-negative elements, and
full rank, and let \bv be a vector with strictly positive elements. Then
$$
\Mdiag{\GM\bv}\minv{\Mdiag{\bv}} \succeq \GM.
$$
(Here $\Mtx{A}\succeq\Mtx{B}$ means that $\Mtx{A} - \Mtx{B}$ is positive
semidefinite.)
\end{lemma}
\begin{proof}
First note that if $\Mtx{A}$ is symmetric, and has non-negative elements,
then $\Mdiag{\Mtx{A}\vone} \succeq \Mtx{A}$ because the matrix
$\Mdiag{\Mtx{A}\vone} - \Mtx{A}$ is symmetric, with non-negative diagonal,
and is diagonally dominant, thus it is positive semidefinite.

Apply that fact with $\Mtx{A}=\Mdiag{\bv}\GM\Mdiag{\bv}$. Then
\begin{align*}
\Mdiag{\Mdiag{\bv}\GM\Mdiag{\bv}\vone} &\succeq \Mdiag{\bv}\GM\Mdiag{\bv},\\
\Mdiag{\bv}\Mdiag{\GM\bv} &\succeq \Mdiag{\bv}\GM\Mdiag{\bv},\\
\Mdiag{\GM\bv} \minv{\Mdiag{\bv}}&\succeq \GM,
\end{align*}
as needed. 
	
Note that the proof relies on strict positivity of the elements
of \bv. By this we can claim a bijection between $\reals{n}$ and
$\Mdiag{\bv}\reals{n}$, which allows us to conclude the last line above.
\end{proof}

\begin{lemma}
Let \GM be a symmetric matrix with non-negative elements, and full rank. 
Let \bv be some vector with non-negative elements such that $\GM\bv$ has strictly
positive elements.
Letting 
$$
\xmp[k+1] \leftarrow \xmp[k] -\Mdiag{\bv}\minv{\Mdiag{\GM\bv}}\wrapParens{\dvp + \GM\xmp[k]},
$$
then $\objf{\xmp[k+1]} \le \objf{\xmp[k]}.$
\label{lemma:pre_ls_update}
\end{lemma}
\begin{proof}
First, consider the case where \bv has strictly positive elements.
Letting
$\GMp[k] = \Mdiag{\GM\bv}\minv{\Mdiag{\bv}},$
the minimum of \objfp[k]{\xmp} occurs at
$-\minv{\GMp[k]}\dvp[k]$, which has value
\begin{equation}
	\begin{split}
		-\minv{\GMp[k]}\dvp[k] &=-\minv{\GMp[k]}\wrapParens{\dvp + \GM\xmp[k] - \GMp[k]\xmp[k]},\\
		&=\xmp[k] - \minv{\GMp[k]}\wrapParens{\dvp + \GM\xmp[k]},\\
		&=\xmp[k] -\Mdiag{\bv}\minv{\Mdiag{\GM\bv}}\wrapParens{\dvp + \GM\xmp[k]} = \xmp[k+1].
	\end{split}
\end{equation}
By \lemmaref{diag_dominance}, 
\objfp[k]{\xmp} dominates \objf{\xmp}, and so $\objf{\xmp[k+1]} \le \objf{\xmp[k]}$.

To prove the theorem for the general case where \bv is simply non-negative,
consider a sequence of strictly positive vectors which converge to \bv, and
apply the argument above, then appeal to continuity of $\objf{\cdot}$.
\end{proof}

\begin{theorem}[\citeauthor{LeeSeung}]
\label{theorem:lee_seung}
Let \GM be a symmetric matrix with non-negative elements, and
full rank. Assume \xmp[k] has non-negative elements, and 
$\GM\xmp[k]$ has strictly positive elements.
If 
\begin{equation}
	\xmp[k+1] \leftarrow - \xmp[k] \hadt \dvp \hadd \wrapParens{\GM\xmp[k]},
\end{equation}
then $\objf{\xmp[k+1]} \le \objf{\xmp[k]}$.
Moreover, the update preserves non-negativity of iterates as long as \dvp has non-positive
elements.
\end{theorem}
\begin{proof}
Let $\bv=\xmp[k]$ in \lemmaref{pre_ls_update}. Then note that
\begin{equation}
	\begin{split}
		\xmp[k+1]
		&=\xmp[k] -\Mdiag{\xmp[k]}\minv{\Mdiag{\GM\xmp[k]}}\wrapParens{\dvp + \GM\xmp[k]},\\
		&= \xmp[k] -\Mdiag{\xmp[k]}\minv{\Mdiag{\GM\xmp[k]}}\dvp - \xmp[k],\\
		&= -\Mdiag{\xmp[k]}\minv{\Mdiag{\GM\xmp[k]}}\dvp,\\
		&=- \xmp[k] \hadt \dvp \hadd \wrapParens{\GM\xmp[k]}.
	\end{split}
\end{equation}
\end{proof}

Returning to the original problem of minimizing the objective of \eqnref{nmf_new_objective}, 
the iterative update is
\begin{multline}
	\XMp[k+1] \leftarrow - \XMp[k] \hadt \wrapParens{\WM{1} - \LM[\top]\WM{0,R}\YM\WM{0,C}\RM[\top]}\hadd \\
	\wrapParens{\LM[\top]\WM{0,R}\LM\XMp[k]\RM\WM{0,C}\RM[\top] + \sum_j\WM{2,R,j}\XMp[k]\WM{2,C,j}}.
\end{multline}

\subsubsection{As a Matrix Factorization Algorithm}
\label{subsec:as_matrix_factorization_algorithm}
Now consider using this iterative update in the solution of \problemref{nmf_def_regularized}.
One starts with initial guesses \LMp[0] and \RMp[0], which are strictly positive, then computes in turn
\begin{align}
	\LMp[1] & \leftarrow - \LMp[0] \hadt \wrapParens{\WM{1,L} - \WM{0,R}\YM\WM{0,C}\trRMp[0]}\hadd\\
	&\phantom{\leftarrow}\quad\nonumber 
	\wrapParens{\WM{0,R}\LMp[0]\RMp[0]\WM{0,C}\trRMp[0] + \sum_j\WM{2,R,L,j}\LMp[0]\WM{2,C,L,j}},\\
	\RMp[1] & \leftarrow - \RMp[0] \hadt \wrapParens{\WM{1,R} - \trLMp[1]\WM{0,R}\YM\WM{0,C}}\hadd\\
	&\phantom{\leftarrow}\quad\nonumber 
	\wrapParens{\trLMp[1]\WM{0,R}\LMp[1]\RMp[0]\WM{0,C} + \sum_j\WM{2,R,R,j}\RMp[0]\WM{2,C,R,j}}.
\end{align}
Then one computes estimates of $\LMp[2], \RMp[2], \LMp[3], \RMp[3], \ldots$

The restriction on the $\WM{1}$ from the symmetric form of the problem translates into the requirements that
\begin{align}
	{\WM{1,L} - \WM{0,R}\YM\WM{0,C}\trRMp[k]} & \le 0,\quad\mbox{and}\\
	{\WM{1,R} - \trLMp[k]\WM{0,R}\YM\WM{0,C}} & \le 0,
\end{align}
elementwise. 
This restriction suffices to keep iterates strictly positive if the initial iterates are also strictly positive.
Note, however, it is not clear how this can be guaranteed \emph{a priori}, as the iterates \RMp[k] and \LMp[k] may become small,
and the \WM{1,\cdot} may be large.
In fact, the \emph{entire point} of $\ell_1$ regularization is to encourage elements of \RMp[k] and \LMp[k] to take value zero.
As a practical matter, then, code which allows arbitrary \WM{1,L}, \WM{1,R} should take the liberty of overriding the input values
and always guaranteeing that the numerators are bounded elementwise by
\begin{equation}
\wrapParens{\WM{1,L} - \WM{0,R}\YM\WM{0,C}\trRMp[k]} \le - \epsilon \ge \wrapParens{\WM{1,R} - \trLMp[k]\WM{0,R}\YM\WM{0,C}}
\end{equation}
for some $\epsilon$. 
This is a compromise for simplicity, and thus this form of the algorithm does not always solve for the input problem.
In the next section we describe a more principled approach that respects the user input.

Again the restriction on the \WM{1} from the symmetric problem cannot be guaranteed \emph{a priori};
instead an implementation must keep the numerators bounded away from zero.
We present the method in \algorithmref{lee_seung_improved}. 
We call it a ``multiplicative update'' since the update steps are elementwise multiply and divide operations
on the initial estimates of \LM and \RM.
For this reason, it has the same limitations as the \citeauthor{LeeSeung} algorithm, namely that once 
an element of \LM or \RM takes value zero, it can never take a non-zero value.
This algorithm assumes the simplified form of \eqnsref{simple_weighting}; the code for the more
general form of the problem is an obvious modification of the update steps.

\begin{algorithm}
	\caption{Multiplicative Update Regularized Non-Negative Matrix Factorization.}
	\label{algorithm:lee_seung_improved}
\begin{algorithmic}
	\Function{MURNMF}{$\YM$, $d$, $\WM{0,R}$, $\WM{0,C}$, $\lambda_{1,L}$, $\lambda_{1,R}$, $\lambda_{2,L}$, $\lambda_{2,R}$, $\gamma_{2,L}$, $\gamma_{2,R}$, $\epsilon > 0$}
\State Initialize random matrices $\LMp[0] > 0$ and $\RMp[0] > 0$ with $d$ columns and rows, respectively.
\State Let $k \gets 0$.
\While {not converged}
	\State Compute numerator $\Mtx{H} \gets \wrapParens{\lambda_{1,L}\mone - \WM{0,R}\YM\WM{0,C}\trRMp[k]}$.
	\State Clip the numerator $\Mtx{H} \gets \Mtx{H} \wedge -\epsilon \mone$.
	\State Let $\FM \gets \wrapParens{\WM{0,R}\LMp[k]\RMp[k]\WM{0,C}\trRMp[k] + \lambda_{2,L}\LMp[k] + \gamma_{2,L}\LMp[k]\wrapParens{\offdiag}}$.
	\State Let $\LMp[k+1] \gets - \LMp[k] \hadt \Mtx{H}\hadd \FM$.
	\State Compute numerator $\Mtx{J} \gets \wrapParens{\lambda_{1,R}\mone - \trLMp[k+1]\WM{0,R}\YM\WM{0,C}}$.
	\State Clip the numerator $\Mtx{J} \gets \Mtx{J} \wedge -\epsilon \mone$.
	\State Let $\FM \gets \wrapParens{\trLMp[k+1]\WM{0,R}\LMp[k+1]\RMp[k]\WM{0,C} + \lambda_{2,R}\RMp[k] + \gamma_{2,R}\RMp[k]\wrapParens{\offdiag}}$.
	\State Let $\RMp[k+1] \gets - \RMp[k] \hadt \Mtx{J}\hadd \FM$.
	\State Increment $k \gets k+1$.
	\State Check convergence.
\EndWhile
	\State \Return $\xmp[k]$
\EndFunction
\end{algorithmic}
\end{algorithm}


\section{Additive Steps and Convergence}
\theoremref{lee_seung} only tells us that the sequence \objf{\xmp[k]} is non-decreasing.
It does \emph{not} guarantee convergence to a global, or even a local, minimum.
In fact, if we only restrict \dv to be non-positive,
it is easy to construct cases where \xmp[k] will not converge
to the constrained optimum.
For if an element of \dv is zero, 
then that same element of \xmp[k] will be zero for all $k > 0$. 
However, the optimum may occur for \xmp with a non-zero value for that element.
(Certainly the \emph{unconstrained} solution $-\minv{\GM}{\dv}$ may have a non-zero 
value for that element.)


Here we analyze the \citeauthor{LeeSeung} update as a more traditional additive step,
rather than a multiplicative step.
Expressing the iteration as an additive step, we have
\begin{equation}
	\xmp[k+1] = \xmp[k] + \hstep[k+1],
\end{equation}
where
\begin{align}
	\hstep[k+1] &= - \xmp[k] \hadt \wrapParens{\dv \hadd \wrapParens{\GM\xmp[k]} + \vone},\\
	&= - \wrapParens{\GM\xmp[k] + \dv} \hadt \xmp[k] \hadd \wrapParens{\GM\xmp[k]},\nonumber\\
	&= - \gradof{\objf{\xmp[k]}} \hadt \xmp[k] \hadd \wrapParens{\GM\xmp[k]}.
\label{eqn:hstep_def}
\end{align}
This may not be the optimal length step in the direction of \hstep[k+1].
If we instead take
\begin{equation}
	\xmp[k+1] = \xmp[k] + c \hstep[k+1],
\end{equation}
then $\objf{\xmp[k+1]}$ is a quadratic function of $c$ with optimum at 
\begin{equation}
	c^{*} = \frac{-\tr{\wrapParens{\GM\xmp[k] + \dv}}\hstep[k+1]}{\qform{\GM}{\hstep[k+1]}}.
\label{eqn:cstar_def}
\end{equation}
However, if $c^{*} > 1$ elements of $\xmp[k] + c^{*} \hstep[k+1]$ could be negative.
One would instead have to select a $c$ that results in a strictly
non-negative $\xmp[k+1]$.
However, if $c^{*} < 1$, then the original algorithm would have overshot the optimum.
\begin{lemma}
Let \GM be a symmetric matrix with non-negative elements, and full rank. 
Let $c^{*}$ be as defined in \eqnref{cstar_def}.
Let $c'$ be the largest number in $\ccinterval{0}{c^{*}}$ such that
all elements of $\xmp[k+1] \leftarrow \xmp[k] + c' \hstep[k+1]$
are non-negative, where $\hstep[k+1]$ is as in \eqnref{hstep_def}.
Then $\objf{\xmp[k+1]} \le \objf{\xmp[k]}$.
\end{lemma}
\begin{proof}
Letting $\funcit{f}{c} = \objf{\xmp[k] + c \hstep[k+1]}$,
$\funcit{f}{c}$ is quadratic in $c$ with positive second derivative.
Then since $c'$ is between $0$ and $c^{*}$, we have $\funcit{f}{c'} \le \funcit{f}{0}$,
which was to be proven.
By construction the elements of $\xmp[k+1]$ are non-negative.
\end{proof}
We note that if $c^{*}\le 1$ then $c'=c^{*}$, since non-negativity is sustained for smaller step sizes.

This iterative update was originally described by \citeauthor{merritt2005interior}
for solution of ``Totally Nonnegative Least Squares'' problems. \cite{merritt2005interior}
Their algorithm has guaranteed convergence 
for minimization of $\normsq{\Mtx{A}\xmp - \vect{b}}$, without the weighting or regularization terms,
under certain conditions. 
They assume strict positivity of $\vect{b}$, which we cannot easily express in terms of \GM and \dv,
and it is not obvious that their proof can be directly used to guarantee convergence.
Without guarantees of convergence, we express their method in our nomenclature as \algorithmref{giqpm}, 
computing the step as $\hstep[k+1] \leftarrow - \gradof{\objf{\xmp[k]}} \hadt \xmp[k] \hadd \wrapParens{\GM\xmp[k]}$.
The vague language around selecting the step length $\tau_k$ is due to \citeauthor{merritt2005interior};
presumably one can choose it randomly, or always take $\tau_k = \wrapParens{\tau + 1}/2$.

%


\subsection{Other Directions}

Imagine, instead, plugging in different values of \bv into the $\xmp[k+1]$ given by \lemmaref{pre_ls_update}.
We claim that setting $\bv = \xmp[k] + \minv{\GM}\dv$
would yield the global unconstrained minimizer for \objf{\xmp}:
\begin{equation}
	\begin{split}
		\xmp[k+1] &\leftarrow 
		\xmp[k] - \Mdiag{\xmp[k] + \minv{\GM}{\dv}}\minv{\Mdiag{\GM\xmp[k] + \dv}}\wrapParens{\dv + \GM\xmp[k]},\\
		&=\xmp[k] - \Mdiag{\xmp[k] + \minv{\GM}{\dv}}\vone,\\
		&= -\minv{\GM}{\dv}.
	\end{split}
\end{equation}

This would seem to be a totally useless computation because we cannot efficiently compute
-\minv{\GM}{\dv}, and it likely violates the positivity constraint.
However, this suggests an alternative iterative step that might have
accelerated convergence.
For example, it might be the case that
setting $\bv = \pospart{\wrapParens{\xmp[k] + \minv{\Mdiag{\vdiag{\GM}}}\dv}},$
might give quicker convergence to a solution, where $\pospart{x}$ is the non-negative part of $x$.

We can also imagine an iterative update based on different descent directions altogether.
For example steepest descent, where $\hstep[k+1] = - \gradof{\objf{\xmp[k]}}$. 
Both of these and the method above can be couched as the general iterative method of
\algorithmref{giqpm}.
We note, however, that steepest descent will fail to make progress in this formulation
when an element of $\xmp[k]$ is zero and the corresponding element of the gradient is positive.
In this case, the step length would be computed as zero and the algorithm would terminate early.
The \citeauthor{LeeSeung} step naturally avoids this problem by scaling the step direction
proportional element-wise to $\xmp[k]$.
It is not clear, however, whether the denominator part of the \citeauthor{LeeSeung} step is necessary,
and perhaps taking steps in the direction of $\hstep[k+1] = - \gradof{\objf{\xmp[k]}} \hadt \xmp[k]$
sufficiently good convergence.

\begin{algorithm}
	\caption{The General Iterative Quadrative Programming Method}
	\label{algorithm:giqpm}
\begin{algorithmic}
\Function{GIQPM\_STEP}{$\xmp[k]$, $\GM$, $\dv$, $\tau \in \oointerval{0}{1}$}
	\State Compute the gradient at \xmp[k]: $\gradof{\objf{\xmp[k]}} \gets {\GM\xmp[k] + \dv}.$
	\State Somehow choose step $\hstep[k+1]$ such that ${\hstep[k+1]}^{\top}\gradof{\objf{\xmp[k]}} < 0$.
	\State Compute maximum allowable step length:\State \quad $\hat{\alpha}_k \gets \min\wrapBraces{\alpha : \xmp[k] + \alpha \hstep[k+1] \ge 0}$.
	\State Compute optimal step length:\State \quad $\alpha^{*}_k \gets - \gradof{\objf{\xmp[k]}}^{\top}\hstep[k+1] / \wrapParens{{\hstep[k+1]}^{\top}\GM\hstep[k+1]}$.
	\State Choose $\tau_k \in \cointerval{\tau}{1}$ and let $\alpha_k$ be the minimum of $\tau_k\hat{\alpha}_k$ and $\alpha^{*}_k$.
	\State Let $\xmp[k+1] \leftarrow \xmp[k] + \alpha_k \hstep[k+1]$.
	\State \Return $\xmp[k+1]$
\EndFunction
	\Function{GIQPM}{$\GM$, $\dv$, $\tau \in \oointerval{0}{1}$}
\State Initialize $\xmp[0] > 0$ and set $k\leftarrow 0$. 
\While {not converged}
	\State Let $\xmp[k+1] \leftarrow\textsc{giqpm\_step}\wrapParens{\xmp[k], \GM, \dv, \tau}$.
	\State Increment $k \leftarrow k+1$.
	\State Check convergence.
\EndWhile
	\State \Return $\xmp[k]$
\EndFunction
\end{algorithmic}
\end{algorithm}

Converting this to the NMF problem as formulated in \subsecref{as_matrix_factorization_algorithm} is a straightforward,
though somewhat tedious, exercise.
We give the algorithm in \algorithmref{lee_seung_improved_II} for the case where the weighting matrices satisfy the 
conditions of Equations \ref{eqn:simple_weighting}.
We note that the restrictions on $\WM{1,R}$ and $\WM{1,L}$ that caused so much headache above are swept under the rug
here with the check for maximum allowable step length, $\hat{\alpha}_k$, in \textsc{giqpm\_step}.

It is not clear \emph{a priori} whether this algorithm converges quickly.
It is very unlikely that \algorithmref{giqpm} is competitive 
for the constrained optimization problem on vectors given in \ref{eqn:cons_quad_opt_problem}. 
Converting to steepest descent is not recommended due to slow convergence, as noted above.
One could swap out the iterative updates of $\LMp[k]$ and $\RMp[k]$ in 
\algorithmref{lee_seung_improved_II} with something like a constrained conjugate gradient.  \cite{li-conjugate-gradient,nocedal2006numerical}
However, one should take several steps of conjugate gradient for each $k$ as the optimization problem changes as we update
the $\LMp[k]$ and $\RMp[k]$.

\begin{algorithm}
	\caption{Additive Update Regularized Non-Negative Matrix Factorization, II.}
	\label{algorithm:lee_seung_improved_II}
\begin{algorithmic}
\Function{GIQPM\_STEP}{$\XMp[k]$, $\gradof{\objf{\XMp[k]}}$, $\Hstep[k+1]$, $\KMp[k+1]$, $\tau_k \in \oointerval{0}{1}$}
	\State Compute maximum allowable step length:\State \quad $\hat{\alpha}_k \gets \min\wrapBraces{\alpha : \XMp[k] + \alpha \HMp[k+1] \ge 0}$.
	\State Compute optimal step length: \State \quad $\alpha^{*}_k \gets - \trace{\tr{\gradof{\objf{\XMp[k]}}} \HMp[k+1]} / \trace{\tr{\Hstep[k+1]}\KMp[k+1]}$. 
	\State \quad (compute these traces without performing matrix multiplies.) 
	\State Let $\alpha_k$ be the minimum of $\tau_k\hat{\alpha}_k$ and $\alpha^{*}_k$.
	\State Let $\XMp[k+1] \leftarrow \XMp[k] + \alpha_k \Hstep[k+1]$.
	\State \Return $\XMp[k+1]$
\EndFunction
\Function{PICK\_DIRECTION}{$\XMp[k]$, $\gradof{\objf{\XMp[k]}}$, $\FM$}
	\State Initialize $\Hstep[k] \gets - \gradof{\objf{\XMp[k]}}\hadt\XMp[k]\hadd\FM$.
	\State Elements of $\Hstep[k]$ for which $\FM$ and $\XMp[k]$ are both zero set to $\fmax{-\gradof{\objf{\XMp[k]}},0}$.
	\State Elements of $\Hstep[k]$ for which $\FM$ is zero but $\XMp[k] > 0$ set to $-\gradof{\objf{\XMp[k]}}\hadt\XMp[k]$.
	\State \Return $\Hstep[k+1]$
\EndFunction
	\Function{AURNMF}{$\YM$, $d$, $\WM{0,R}$, $\WM{0,C}$, $\lambda_{1,L}$, $\lambda_{1,R}$, $\lambda_{2,L}$, $\lambda_{2,R}$, $\gamma_{2,L}$, $\gamma_{2,R}$, $\tau \in \oointerval{0}{1}$}
\State Initialize random $\LMp[0] > 0$ with $d$ columns and $\RMp[0] > 0$ with $d$ rows.
\State Let $k \leftarrow 0$.
\While {not converged}
	\State Choose $\tau_k \in \cointerval{\tau}{1}$.
	\State $\DM \gets \wrapParens{\lambda_{1,L}\mone - \WM{0,R}\YM\WM{0,C}\trRMp[k]}$.  
	\State $\FM \gets {\WM{0,R}\LMp[k]\qoform{\WM{0,C}}{\RMp[k]} + \lambda_{2,L} \LMp[k] + \gamma_{2,L}\LMp[k]\wrapParens{\offdiag}}$ 
	\State $\gradof{\objf{\LMp[k]}} \gets \FM + \DM$. 
	\State $\Hstep[k+1] \gets \textsc{pick\_direction}\wrapParens{\LMp[k], \gradof{\objf{\LMp[k]}}, \FM}.$
	\State $\KMp[k+1] \gets \WM{0,R}\Hstep[k+1]\qoform{\WM{0,C}}{\RMp[k]} + \lambda_{2,L}\Hstep[k+1]$. 
	\State Let $\LMp[k+1] \gets \textsc{giqpm\_step}\wrapParens{\LMp[k], \gradof{\objf{\LMp[k]}}, \Hstep[k+1], \KMp[k+1], \tau_k}$.
	\State $\DM \gets \wrapParens{\lambda_{1,R}\mone - \trLMp[k+1]\WM{0,R}\YM\WM{0,C}}$.  
	\State $\FM \gets {\qform{\WM{0,R}}{\LMp[k+1]}\RMp[k]\WM{0,C} + \lambda_{2,R} \RMp[k] + \gamma_{2,R}\RMp[k]\wrapParens{\offdiag}}$ 
	\State $\gradof{\objf{\RMp[k]}} \gets \FM + \DM$. 
	\State $\Hstep[k+1] \gets \textsc{pick\_direction}\wrapParens{\RMp[k], \gradof{\objf{\RMp[k]}}, \FM}$.
	\State $\KMp[k+1] \gets {\qform{\WM{0,R}}{\LMp[k+1]}\Hstep[k+1]\WM{0,C} + \lambda_{2,R} \Hstep[k+1]}$ 
	\State Let $\RMp[k+1] \gets \textsc{giqpm\_step}\wrapParens{\RMp[k], \gradof{\objf{\RMp[k]}}, \Hstep[k+1], \KMp[k+1], \tau_k}$.
	\State Increment $k \leftarrow k+1$.
	\State Check convergence.
\EndWhile
	\State \Return $\xmp[k]$
\EndFunction
\end{algorithmic}
\end{algorithm}

\section{Simulations}

The multiplicative and additive algorithms are implemented in the R package \textsc{rnnmf} written by the author.  \cite{rnnmfrnnmf-Manual}
Here we briefly summarize the results of using the code on generated data.
We first generate a $\YM$ the exactly equals $\LM\RM$ for some non-negative uniformly random $\LM$ and $\RM$.
In the first simulations we take $\LM$ to be \bby{30}{2}, and $\RM$ to be
\bby{2}{8}.
We randomly generate starting iterates, 
\bby{30}{2} matrix \LMp[0] and
\bby{2}{8} matrix \RMp[0].
We use the same starting point for both the additive and multiplicative algorithms.
We compute the Frobenius norm of the error, which is to say 
$$\sqrt{\trace{\gram{\wrapParens{\YM - \LMp[k]\RMp[k]}}}},$$
for each iteration.

In \figref{mc_sims_plot1} we plot the error versus step for the additive and multiplicative methods.
The additive method uses the optimal step size in the chosen direction.
While not shown here, choosing the naive step without optimizing converges slower in iterations but may ultimately be faster computationally.
The additive method outperforms the multiplicative method in convergence per step.
Convergence of both is ``lumpy'', with apparent phase changes where, we suspect, different directions become the dominant source of approximation error,
and then are smoothed out.

\begin{knitrout}\small
\definecolor{shadecolor}{rgb}{0.969, 0.969, 0.969}\color{fgcolor}\begin{figure}[h]
\includegraphics[width=0.975\textwidth,height=0.691\textwidth]{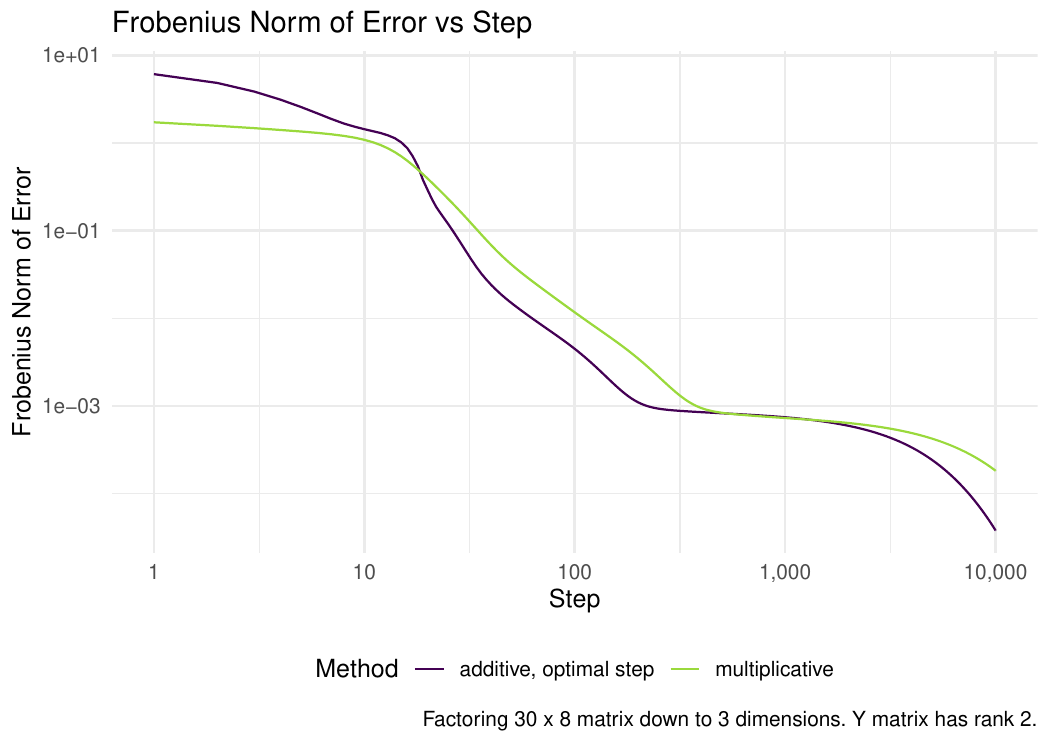} \caption[The Frobenius norm is plotted versus step for two methods for a small problem]{The Frobenius norm is plotted versus step for two methods for a small problem.}\label{fig:mc_sims_plot1}
\end{figure}

\end{knitrout}

We then apply the code to another test case.
Here we generate $\LM$ to be \bby{40}{3}, and $\RM$ to be
\bby{3}{10}.
Again we test both the additive and multiplicative algorithms, but test the effect of sparse starting iterates.
We randomly generate starting iterates, 
\bby{40}{4} matrix \LMp[0] and
\bby{4}{10} matrix \RMp[0].
First we do this where about $\frac{1}{3}$ of the elements of $\LMp[0]$ and $\RMp[0]$ are zero, testing both algorithms.
We repeat the experiment for the same \YM, but generate strictly positive
$\LMp[0]$ and $\RMp[0]$.

In \figref{mc_sims_plot2} we plot the error versus step for both methods and both choices of starting iterates.
Again we see somewhat slower convergence for the multiplicative method, at least measured in iterates.
Moreover, the multiplicative method fails to converge for the sparse starting iterates. 
As we noted above, the multiplicative method cannot change an element from zero to non-negative,
thus the lack of convergence for sparse initial estimates is not surprising.

\begin{knitrout}\small
\definecolor{shadecolor}{rgb}{0.969, 0.969, 0.969}\color{fgcolor}\begin{figure}[h]
\includegraphics[width=0.975\textwidth,height=0.691\textwidth]{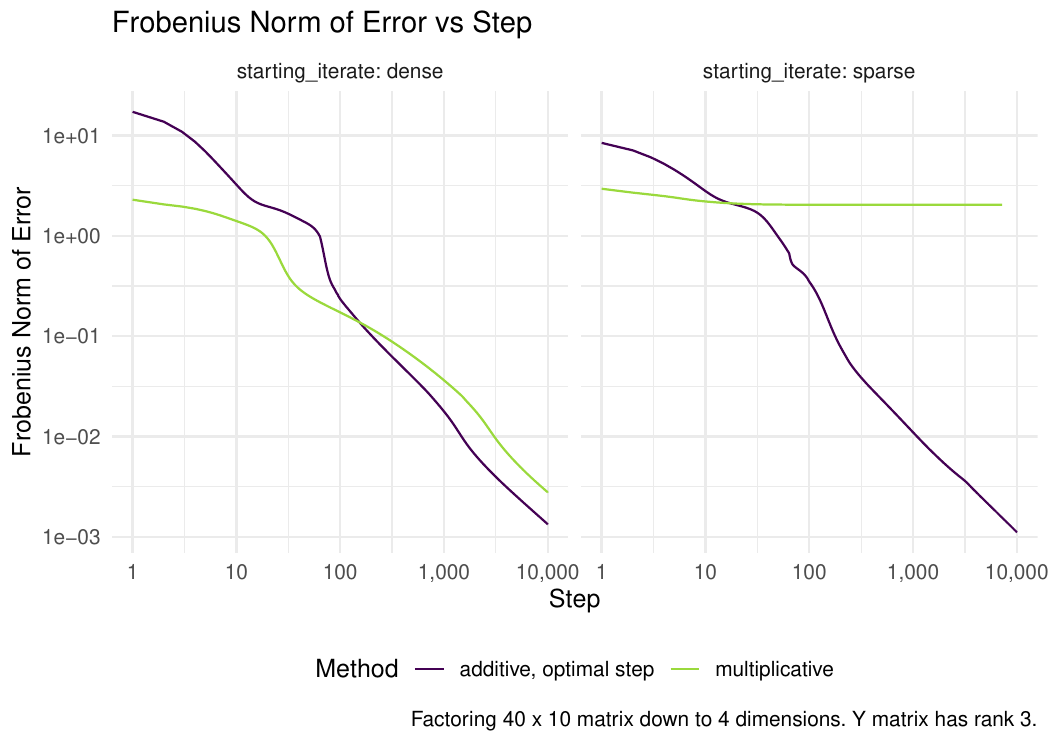} \caption[The Frobenius norm is plotted versus step for two methods for a small problem]{The Frobenius norm is plotted versus step for two methods for a small problem. Starting iterates are taken to be sparse or dense.}\label{fig:mc_sims_plot2}
\end{figure}

\end{knitrout}


\section{Applications}

One straightforward application of non-negative matrix factorization is in chemometrics \cite{pav_nmf},
where one observes the absorption spectra of some samples at multiple wavelengths of light.
The task is to approximate the observed absorption as the product of some non-negative combination
of some non-negative absorption spectra of the underlying analytes.
When the underlying analytes have known absorption spectra the task is like a regression;
when they do not, it is more like a dimensionality reduction or unsupervised learning problem.

We consider a more unorthodox application here, namely an analysis of cocktail recipes.
From a reductionist point of view, a cocktail is a mixture of a number of constituent liquid ingredients.
We use a database of 5,778 cocktail recipes, 
scraped from the website \url{www.kindredcocktails.com},
and distributed in an R package.  \cite{cocktailApp-Manual}
We filter down this selection, removing ingredients which appear in fewer than 
10 cocktails, then removing cocktails with those ingredients.
This winnows the dataset down to 3,814 cocktails.
The upstream website allows visitors to cast votes on cocktails.
We use the number of votes to filter the cocktails.
The median cocktail, however, has only 2 votes.
We remove cocktails with fewer than
2 votes, resulting in a final set of 2,405 cocktails.

We express the cocktails as a sparse non-negative 
\bby{2405}{280} matrix \YM
whose entry at $i,j$ indicates the proportion of cocktail $i$ that is ingredient $j$.
We have normalized the matrix \YM to have rows that sum to 1.
We then performed a basic non-negative matrix factorization $\YM \approx \LM\RM$, without regularization.
We do this for rank $d$ approximation, meanining $\LM$ has $d$ columns and $\RM$ has $d$ rows.
To estimate the quality of the approximation we compute an $R^2$ value.
This is computed as
\begin{equation}
R^2 = 1 - \frac{\trace{\gram{\wrapParens{\YM - \LMp[k]\RMp[k]}}}}{\trace{\gram{\wrapParens{\YM - \vone\RM[0]}}}}.
\end{equation}
Here $\RM[0]$ is the vector of the column means of \YM.
One can think of $\RM[0]$ as encoding the ``Jersey Turnpike'', a drink with a small amount of \emph{every} ingredient.

\begin{knitrout}\small
\definecolor{shadecolor}{rgb}{0.969, 0.969, 0.969}\color{fgcolor}\begin{figure}[h]
\includegraphics[width=0.975\textwidth,height=0.691\textwidth]{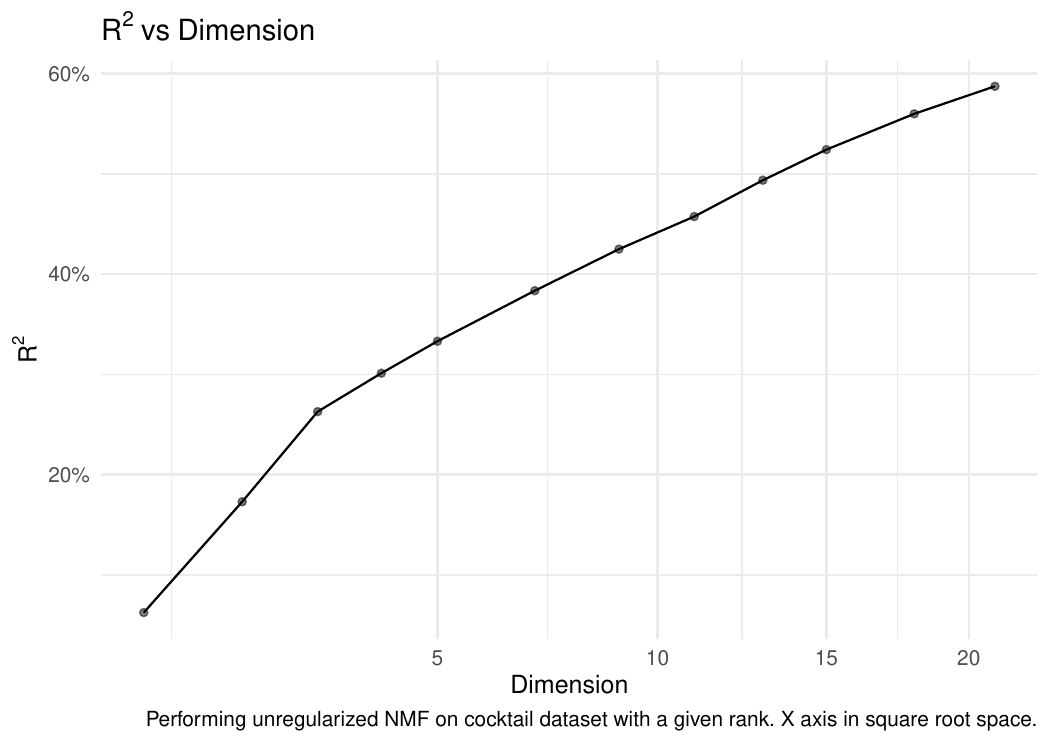} \caption[$R^2$ is plotted versus $d$ for the unregularized factorization of the cocktail data]{$R^2$ is plotted versus $d$ for the unregularized factorization of the cocktail data.}\label{fig:coapp_R2_plot}
\end{figure}

\end{knitrout}

We plot $R^2$ versus $d$ in \figref{coapp_R2_plot}, with the $x$ axis in square root space,
giving a near-linear fit but with a crook, or knee, around $d=3$, which has an $R^2 \approx 26\%.$
We will consider factorizations with this $d$, and also with $d=9$ for which 
$R^2 \approx 42\%.$
To account for the differential popularity of different cocktails,
we use the number of votes to weight the error of our approximation.
We take $\WM{0,R}$ to be the diagonal matrix consisting of the number of votes for each cocktail.
We perform factorization under this weighted objective.

\begin{table}

\caption{\label{tab:coapp_first_table}The proportions of ingredients in the 3 latent cocktails from the weighted fit are shown. Ingredients which are less than 0.03 of the total are not shown.}
\centering
\begin{tabular}[t]{l|l|r}
\hline
Latent Cocktail & Ingredient & Proportion\\
\hline
factor1 & Gin & 0.433\\
\hline
factor1 & Other (256 others) & 0.415\\
\hline
factor1 & Lemon Juice & 0.067\\
\hline
factor1 & Sweet Vermouth & 0.046\\
\hline
factor1 & Lime Juice & 0.038\\
\hline
factor2 & Bourbon & 0.474\\
\hline
factor2 & Other (243 others) & 0.350\\
\hline
factor2 & Sweet Vermouth & 0.071\\
\hline
factor2 & Lemon Juice & 0.036\\
\hline
factor2 & Campari & 0.035\\
\hline
factor2 & Cynar & 0.034\\
\hline
factor3 & Rye & 0.490\\
\hline
factor3 & Other (234 others) & 0.408\\
\hline
factor3 & Sweet Vermouth & 0.102\\
\hline
\end{tabular}
\end{table}

\begin{knitrout}\small
\definecolor{shadecolor}{rgb}{0.969, 0.969, 0.969}\color{fgcolor}\begin{figure}[h]
\includegraphics[width=0.975\textwidth,height=0.5571\textwidth]{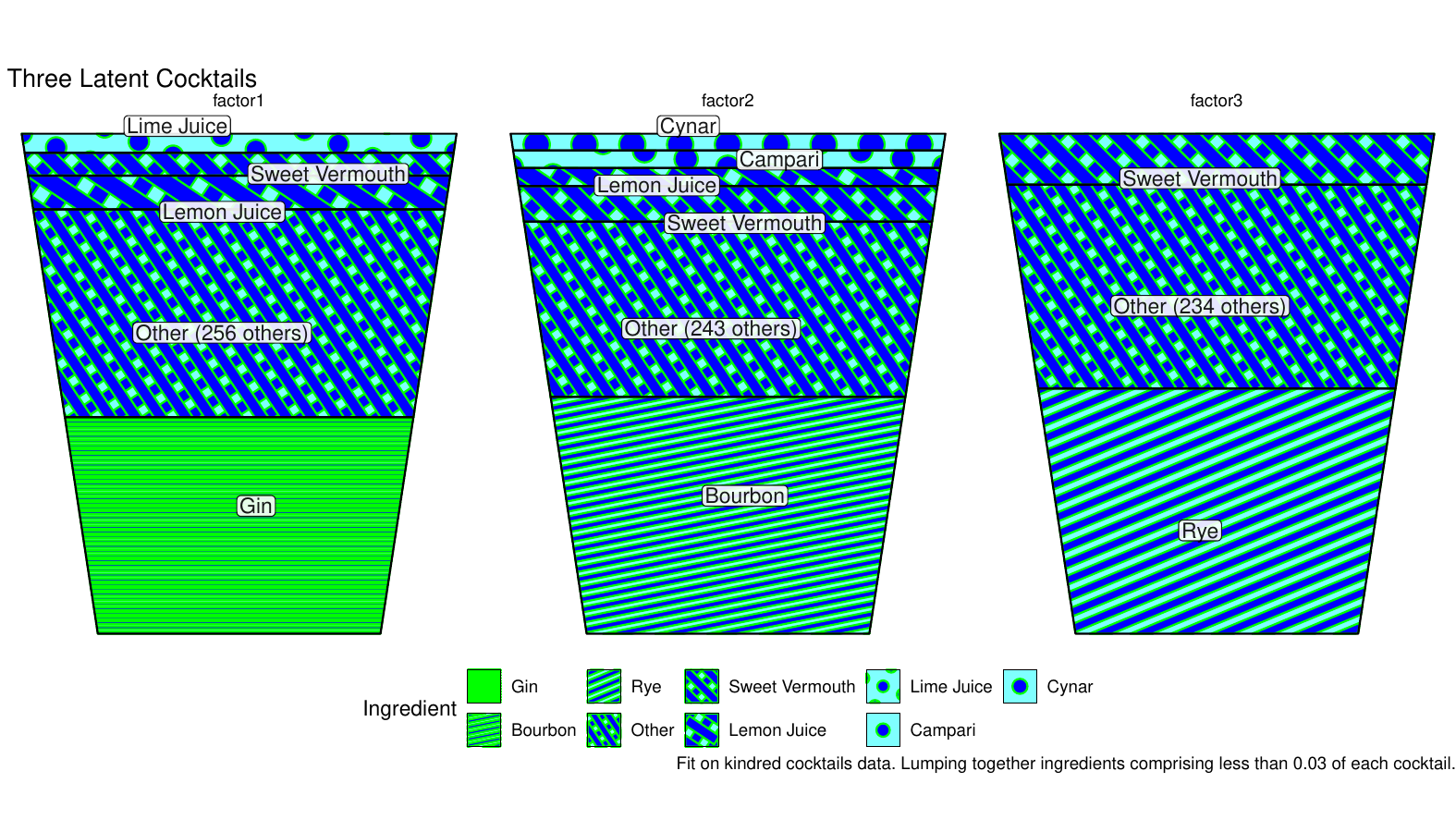} \caption{The first three latent cocktails are shown. Ingredients making less than 0.03 of the total are lumped together.  Patterns are courtesy of the \textsc{ggpattern} package.  \cite{ggpattern}}\label{fig:coapp_first_plot}
\end{figure}

\end{knitrout}
Before displaying the results, we note that a matrix factorization is only determined up to 
rescaling and rotation.
That is, if \Mtx{Q} is an invertible matrix, then the following two factorizations
are equivalent:
$$
\YM \approx \LM\RM = \wrapParens{\LM\Mtx{Q}} \wrapParens{\minv{\Mtx{Q}}\RM}.
$$
These two factorizations might not be equivalent for the regularized factorization objective, however.
We make use of this equivalence when demonstrating the output of our algorithm:
we use matrix \Mtx{Q} such that $\RM$ has unit row sums, and we rearrange the order of the factors
such that the column sums of $\LM$ are non-increasing. 
With that in mind, we display the first three latent factor cocktails in
in \tabref{coapp_first_table}, lumping together those ingredients which make up less than $0.03$ of the factor cocktails.

While the table is definitive, it is hard to appreciate the distribution of ingredients in the latent factor cocktails.
Towards that end, we plot them in stacked barcharts in \figref{coapp_first_plot}.
These barcharts are styled to look like drinks, what one might call ``tumbler plots''.
We notice that the latent factor cocktails contain a large number of ingredients falling under the display limit of 0.03.
We also see repetition of ingredients in the factor cocktails, with Sweet Vermouth appearing in appreciable amounts in all three.

To remedy that, we fit again, this time using 
$\lambda_{1,L}=\lambda_{1,R}=0.4$ and 
$\gamma_{2,R} = 0.25$.
It is important to note that because we decompose 
$\YM \approx \LM\RM$ that applying an $\ell_1$ penalty solely on 
$\RM$ could result in a rescaling of output such that elements of $\RM$ are
small in magnitude and elements of $\LM$ are large.
That is, because 
$\LM\RM = \wrapParens{\LM\Mtx{Q}} \wrapParens{\minv{\Mtx{Q}}\RM}$,
we could take $\Mtx{Q} = v \eye$ for large $v$ and make the $\ell_1$ penalty
on \RM irrelevant without really making the decomposition sparse.
To address this, one mst typically apply an $\ell_1$ penalty to both \LM and \RM as we do here.

We chose the values of $\lambda_{1,L}, \lambda_{1,R}, \gamma_{2,R}$ by trial and error,
as unfortunately there is no theory to guide us here. 
Forging on, we plot the results in \figref{coapp_second_plot}.
We see there that the factor cocktails are now contain far fewer ingredients, though they all
hew to the formula of base spirit plus vermouth, with some extras.
These seem more ``spirit-forward'' than a typical cocktail, but \emph{de gustibus non est disputandum}.

\begin{knitrout}\small
\definecolor{shadecolor}{rgb}{0.969, 0.969, 0.969}\color{fgcolor}\begin{figure}[h]
\includegraphics[width=0.975\textwidth,height=0.5571\textwidth]{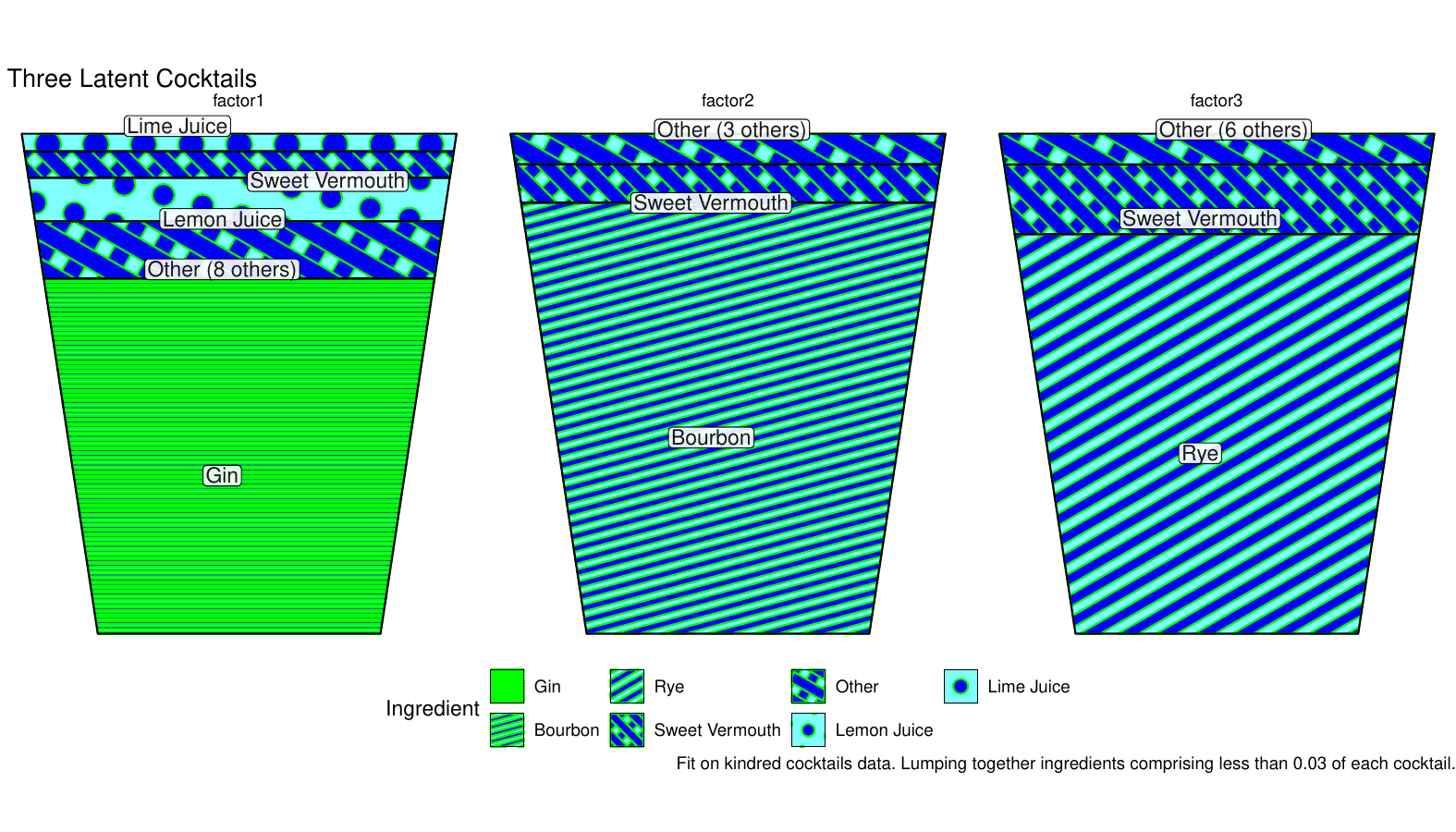} \caption[The second fit of three latent cocktails are shown]{The second fit of three latent cocktails are shown. Ingredients making less than 0.03 of the total lumped together.}\label{fig:coapp_second_plot}
\end{figure}

\end{knitrout}

Finally, we perform another fit, this time with $d=9$,
and using 
$\lambda_{1,L}=\lambda_{1,R}=0.5$,
$\lambda_{2,L}=\lambda_{2,R}=2.5$,
$\gamma_{2,R} = 0.25$, values again chosen by trial and error.
We plot the results of this fit in \figref{coapp_third_plot}.
Again we see heavy use of base spirits with a sprinkling of secondary ingredients.
One can imagine in the limit as $d$ approaches the number of base ingredients
that the factorization could collapse to the trivial one $\YM = \YM \eye$.
We see that behavior here with the top nine most popular ingredients 
dominating the latent factor cocktails.
Thus it is hard to view non-negative matrix factorization as a good way of
quantifying and describing the space of cocktails.
Probably a better approach would be to recognize that cocktails evolve
from other cocktails, then try to infer and describe the evolutionary structure.

\begin{knitrout}\small
\definecolor{shadecolor}{rgb}{0.969, 0.969, 0.969}\color{fgcolor}\begin{figure}[h]
\includegraphics[width=0.975\textwidth,height=1.365\textwidth]{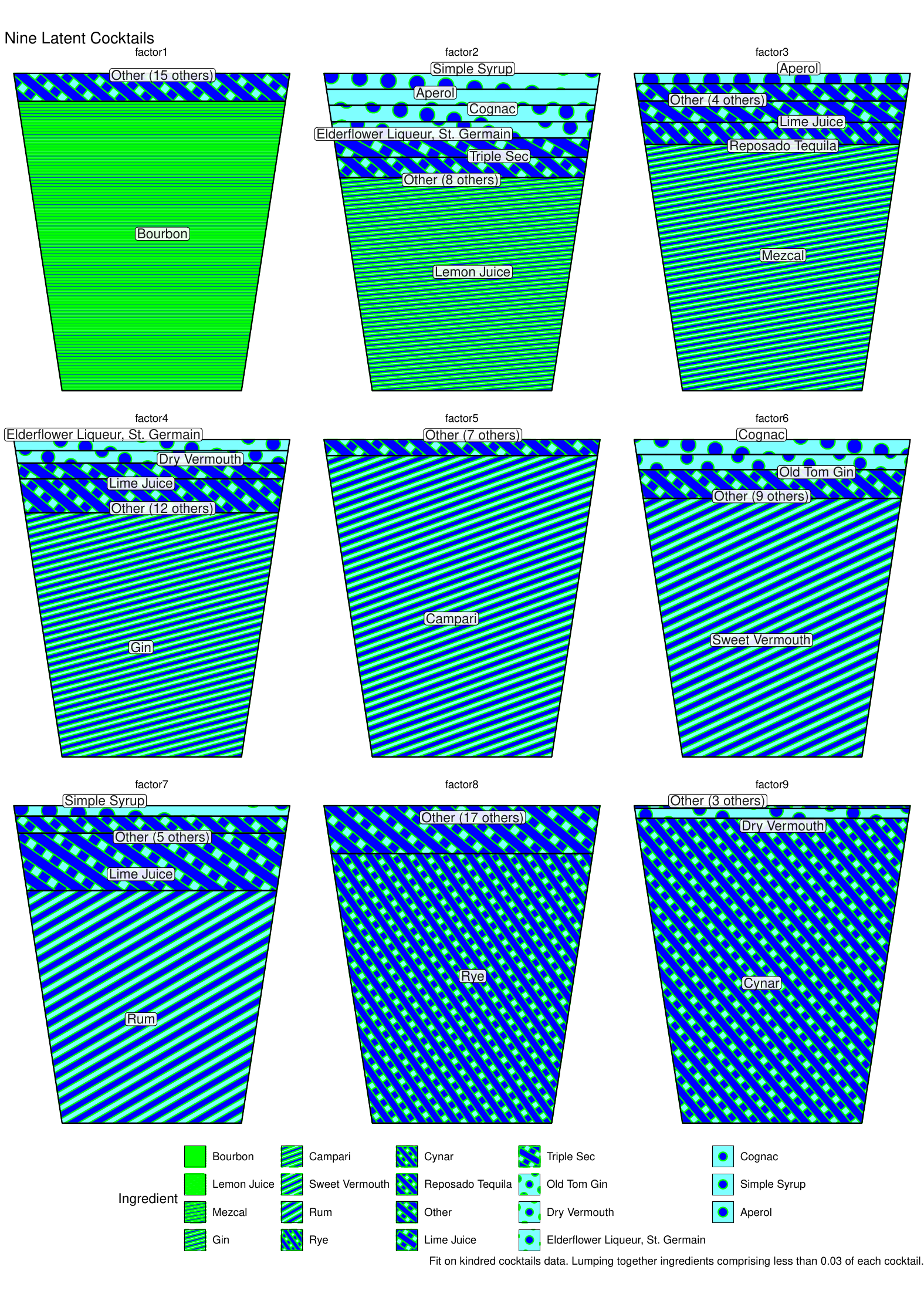} \caption[The nine latent cocktails are shown]{The nine latent cocktails are shown. Ingredients making less than 0.03 of the total lumped together.}\label{fig:coapp_third_plot}
\end{figure}

\end{knitrout}


\bibliographystyle{plainnat}
\bibliography{common,rauto}

\appendix

\section{Matrix Identities}
\label{sec:matrix_identities}

The following identities are useful for switching between matrix and vectorized representations.
\cite{schacke2013kronecker}
\begin{align}
\fvec{\Mtx{B}\Mtx{C}\Mtx{D}} &= \wrapParens{\AkronB{\tr{\Mtx{D}}}{\Mtx{B}}}\fvec{\Mtx{C}}.\\
\trace{\tr{\Mtx{A}}\Mtx{C}} &= \tr{\fvec{\Mtx{A}}}\fvec{\Mtx{C}}.\\
\trace{\tr{\Mtx{A}}\Mtx{B}\Mtx{C}\Mtx{D}} &= \tr{\fvec{\Mtx{A}}}\wrapParens{\AkronB{\tr{\Mtx{D}}}{\Mtx{B}}}\fvec{\Mtx{C}}.\\
\trace{\tr{\Mtx{A}}\Mtx{B}\Mtx{A}\Mtx{D}} &= \normsq[{\AkronB{\tr{\Mtx{D}}}{\Mtx{B}}}]{\fvec{\Mtx{A}}}.\\
\tr{\wrapParens{\AkronB{\Mtx{A}}{\Mtx{B}}}} &= \AkronB{\tr{\Mtx{A}}}{\tr{\Mtx{B}}}.
\end{align}


\end{document}